%% file: main.tex
\documentclass[sigconf]{acmart}

\usepackage{smile}
\usepackage[ruled,linesnumbered]{algorithm2e}
\usepackage{balance}

\newcommand{\model}{GG-ODE}
\newcommand{\te}{t,e}

\AtBeginDocument{%
  \providecommand\BibTeX{{%
    \normalfont B\kern-0.5em{\scshape i\kern-0.25em b}\kern-0.8em\TeX}}}

\copyrightyear{2023}
\acmYear{2023}
\setcopyright{rightsretained}
\acmConference[KDD '23]{Proceedings of the 29th ACM SIGKDD Conference on Knowledge Discovery and Data Mining}{August 6--10, 2023}{Long Beach, CA, USA} \acmBooktitle{Proceedings of the 29th ACM SIGKDD Conference on Knowledge Discovery and Data Mining (KDD '23), August 6--10, 2023, Long Beach, CA, USA} \acmDOI{10.1145/3580305.3599362} \acmISBN{979-8-4007-0103-0/23/08}

\ccsdesc[500]{Information systems~Physical data models}
\ccsdesc[500]{Computing methodologies~Spatial and physical reasoning}

\begin{document}

\title{Generalizing Graph ODE for Learning
Complex System Dynamics across Environments 
}

\author{Zijie Huang}
\affiliation{%
  \institution{University of California, Los Angeles}
  \city{Los Angeles}
  \country{CA}}
\email{zijiehuang@cs.ucla.edu}

\author{Yizhou Sun}
\affiliation{%
  \institution{University of California, Los Angeles}
  \city{Los Angeles}
  \country{CA}
}
\email{yzsun@cs.ucla.edu}

\author{Wei Wang}
\affiliation{%
  \institution{University of California, Los Angeles}
  \city{Los Angeles}
  \country{CA}
}
\email{weiwang@cs.ucla.edu}

\begin{abstract}

Learning multi-agent system dynamics has been extensively studied for various real-world applications, such as molecular dynamics in biology, 
multi-body system in physics, and particle dynamics in material science. 
Most of the existing models are built to learn single system dynamics, which learn the dynamics from observed historical data and predict the future trajectory. 
In practice, however, we might observe multiple systems that are generated across different environments, which differ in latent exogenous factors such as temperature and gravity.
One simple solution is to learn multiple environment-specific models, but it fails to exploit the potential commonalities among the dynamics across environments and offers poor prediction results where per-environment data is sparse or limited.
Here, we present \model~(\textbf{\underline{G}}eneralized \textbf{\underline{G}}raph \textbf{\underline{O}}rdinary \textbf{\underline{D}}ifferential \textbf{\underline{E}}quations), a machine learning framework for learning continuous multi-agent system dynamics across environments. Our model learns system dynamics using neural ordinary differential equations (ODE) parameterized by Graph Neural Networks (GNNs) to capture the continuous interaction among agents. We achieve the model generalization by assuming 
the dynamics across different environments are governed by common physics laws that can be captured via learning a shared ODE function. The distinct latent exogenous factors learned for each environment are incorporated into the ODE function to account for their differences. To improve model performance, we additionally design two regularization losses to (1) enforce the orthogonality between the learned initial states and exogenous factors via mutual information minimization; and (2) reduce the temporal variance of learned exogenous factors 
within the same system via contrastive learning. Experiments over various physical
simulations show that our model can accurately predict system dynamics, especially in the long range, and can generalize well to new systems with few observations.

\end{abstract}


\keywords{Graph Neural Networks; Neural ODE; Dynamical Systems; Representation Learning}

\maketitle

\input{01-Introduction.tex}
\input{02-Problem.tex}
\input{03-RelatedWork.tex}

\input{04-Method.tex}

\input{05-Experiment.tex}
\input{06-Conclusion.tex}

\begin{acks}
This work was partially supported by NSF 1829071, 2031187, 2106859, 2119643, 2200274, 2211557, 1937599, 2303037, NASA, research awards from Amazon, Cisco, NEC, and DARPA \#HR00112290103, DARPA \#HR0011260656. We would like to thank Mathieu Bauchy, Han Liu and Abhijeet Gangan for their help to the dataset generation procedure and valuable discussion throughout the project.
\end{acks}

\bibliographystyle{ACM-Reference-Format}
\balance
\bibliography{reference}

\newpage
\appendix
\input{07-Appendix.tex}

\end{document}

%% file: 01-Introduction.tex
\section{Introduction}
Building a simulator that can understand and predict multi-agent system dynamics is a crucial research topic spanning over a variety of domains such as planning and control in robotics~\cite{ICRA}, where the goal is to generate future trajectories of agents based on what has been seen in the past. Traditional simulators can be very expensive to create and use~\cite{jure} as it requires sufficient domain knowledge and tremendous computational resources to generate high-quality results\footnote{To date, out of the 10 most powerful supercomputers in the world, 9 of them are used for simulations, spanning the fields of cosmology, geophysics and fluid dynamics~\cite{jure_9}}. Therefore, learning a neural-based simulator directly from data that can approximate the behavior of traditional simulators becomes an attractive alternative.

As the trajectories of agents are usually coupled with each other and co-evolve along with the time, existing studies on learning system dynamics from data usually view the system as a graph and employ Graph Neural Networks (GNNs) to approximate pair-wise node (agent) interaction to impose strong inductive bias~\cite{IN}. As a pioneering work, Interaction Networks (IN)~\cite{IN} decompose the system into distinct objects and relations, and learn to reason about the consequences of their interactions and dynamics. Later work incorporates domain knowledge~\cite{li2019propagation}, graph structure variances~\cite{pfaff2020learning}, 
 and equivariant representation learning~\cite{shi2021learning,gasteiger2021gemnet} into learning from discrete GNNs, achieving state-of-the-art performance in various domains including mesh-based physical simulation~\cite{pfaff2020learning} and molecular prediction~\cite{gao2022abinitio}.
 However, these $\textit{discrete}$ models usually suffer from low accuracy in long-range predictions as (1) they approximate the system by discretizing observations into some fixed timestamps and are trained to make a single forward-step prediction and (2) their discrete nature fails to adequately capture systems that are continuous in nature such as the spread of COVID-19~\cite{CG-ODE} and the movements of an n-body system~\cite{NÅÅÅRI,LG-ODE}.

Recently, researchers propose to combine ordinary differential equations (ODEs) - the principled way for modeling dynamical systems in a $\textit{continuous}$ manner in the past, with GNNs to learn continuous-time dynamics on complex networks in a data-driven way~\cite{LG-ODE,CG-ODE,ndcn,hope,CFG-ODE}. These Graph-ODE methods have demonstrated the power of capturing long-range dynamics, and are capable of learning from irregular-sampled partial observations~\cite{LG-ODE}. They usually assume all the data are generated from one single system, and the goal is to learn the system dynamics from historical trajectories to predict the future.
In practice, however, we might observe data that are generated from multiple systems, which can differ in their environments. For example, we may observe particle trajectories from systems that are with different temperatures, which we call exogenous factors.
These exogenous factors can span over a wide range of settings such as particle mass, gravity, and temperature ~\cite{jure,latent_causual,confounder} across environments. One simple solution is to learn multiple environment-specific models, but it can fail to exploit the potential commonalities across environments and make accurate predictions for environments with sparse or zero observations. In many useful contexts, the dynamics in multiple environments share some similarities, yet being distinct reflected by the (substantial) differences in the observed trajectories. For example, considering the movements of water particles within multiple containers of varying shapes, the trajectories are driven by both the shared pair-wise physical interaction among particles (i.e. fluid dynamics) and the different shapes of the containers where collisions can happen when particles hit the boundaries. Also, the computational cost for training multiple environment-specific models would be huge. More challengingly, the exogenous factors within each environment can be latent, such as we only know the water trajectories are from different containers, without knowing the exact shape for each of them. Therefore, how to learn a single efficient model that can generalize across environments by considering both their commonalities and the distinct effect of per-environment latent exogenous factors remains unsolved. This model, if developed, may help us predict dynamics for systems under new environments with very few observed trajectories.

Inspired by these observations, in this paper, we propose Generalized Graph ODE (\model), a general-purpose continuous neural simulator that learns multi-agent system dynamics across environments. Our key idea is to assume the dynamics across environments are governed by common physics laws that can be captured via learning a shared ODE function. We introduce in the ODE function a learnable vector representing the distinct latent exogenous factors for each environment to account for their differences. We learn the representations for the latent exogenous factors from systems' historical trajectories through an encoder by optimizing the prediction goal.  In this way, different environments share the same ODE function framework while incorporating environment-specific factors in the ODE function to distinguish them. 

However, there are two main challenges in learning such latent exogenous factor representations. Firstly, since both the latent initial states for agents and the latent exogenous factors are learned through the historical trajectory data, how can we differentiate them to guarantee they have different semantic meanings? Secondly, when inferring from different time windows from the same trajectory, how can we guarantee the learned exogenous factors are for the same environment? 

Towards the first challenge, we enforce the orthogonality between the initial state encoder and the exogenous factor encoder via mutual information minimization. For the second challenge, we reduce the variance of learned exogenous factors within the same environment via a contrastive learning loss. We train our model in a multi-task learning paradigm where we mix the training data from multiple systems with different environments. In this way, the model is expected to fast adapt to other unseen systems with a few data points. We conduct extensive experiments over a wide range of physical systems, which show that our \model~is able to accurately predict system dynamics, especially in the long range.

The main contributions of this paper are summarized as follows: 

\begin{itemize}
    \item We investigate the problem of learning continuous multi-agent system dynamics across environments. We propose a novel framework, known as \model, which describes the dynamics for each system with a shared ODE function and an environment-specific vector for the latent exogenous factors to capture the commonalities and discrepancies across environments respectively.
    \item We design two regularization losses to guide the learning process of the latent exogenous factors, which is crucial for making precise predictions in the future.
    \item Extensive experiments verify the effectiveness of GG-ODE to accurately predict system dynamics, especially in the long range prediction tasks. \model~also generalizes well to unseen or low-resource systems that have very few training samples.
\end{itemize}  

%% file: 02-Problem.tex
\section{Problem Definition}\label{sec:problem}
We aim to build a neural simulator to learn continuous multi-agent system dynamics automatically from data that can be generalized across environments. Throughout this paper, we use boldface uppercase letters to denote matrices or vectors, and regular lowercase letters to represent the values of variables.

We consider a multi-agent dynamical system of $N$ interacting agents as an evolving interaction graph $\mathcal{G}^t = \{\mathcal{V},\mathcal{E}^t\}$, where nodes are agents and edges are interactions between agents that can change over time. For each dynamical system, we denote $e\in E$ as the environment from which the data is acquired. We denote $\bm{X}^{t,e} \in \mathcal{X}$ as the feature matrix for all $N$ agents and $\bm{x}_i^{t,e}$ as the feature vector of agent $i$ at time $t$ under environment $e$. The edges between agents are assigned if two agents are within a connectivity radius $R$ based on their current locations $\bm{p}_i^{\te}$ which is part of the node feature vector, i.e. $\bm{p}_i^{\te}\in \bm{x}_i^{\te}$. They reflect the local interactions of agents and the radius is kept constant over time~\cite{jure}.

Our model input consists of the trajectories of $N$ agents over $K$ timestamps $X^{t_{1:K},e}=\{\bm{X}^{t_1,e}, \bm{X}^{t_2,e}, \ldots, \bm{X}^{t_K,e}\}$, where the timestamps $t_1,t_2\cdots t_K$ can have non-uniform intervals and be of any continuous values. Our goal is to learn a generalized simulator $s_{\theta}:X^{t_{1:K},e}\rightarrow Y^{t_{K+1:T},e}$ that predicts node dynamics in the future for any environment $e$. 
Here $\bm{Y}^{\te}\in\mathcal{Y}$ represents the targeted node dynamic information at time $t$, and can be a subset of the input features. We use $\bm{y}_i^{\te}$ to denote the targeted node dynamic vector of agent $i$ at time $t$ under environment $e$. 

%% file: 03-RelatedWork.tex
\section{Preliminaries and Related Work}

\subsection{Dynamical System Simulations with Graph Neural Networks (GNNs)}
Graph Neural Networks (GNNs) are a class of neural networks that operate on graph-structured data by passing local messages\cite{GCN,GAT,GIN,CG-ODE,SSAGA}.
They have been extensively employed in various applications such as node classification~\cite{wu2020unsupervised,zhao2021graphsmote}, link prediction~\cite{qu2020continuous,benson2019link}, and recommendation systems~\cite{he2020lightgcn,wang2019neural,didiff,hashtag}.
By viewing each agent as a node and interaction among agents as edges, GNNs have shown to be efficient for approximating pair-wise node interactions and achieved accurate predictions for multi-agent dynamical systems \cite{NRI,compositional,jure}. 
The majority of existing studies propose discrete GNN-based simulators where they take the node features at time $t$ as input to predict the node features at time $t$+1. To further capture the long-term temporal dependency for predicting future trajectories, some work utilizes recurrent neural networks such as RNN, LSTM or self-attention mechanism to make prediction at time $t$ +1 based on the historical trajectory sequence within a time window~\cite{VGRNN,Dygnn,co-evolve,hgt}. However, they all restrict themselves to learn a one-step state transition function. Therefore, when successively apply these one-step simulators to previous predictions in order to generate the rollout trajectories, error accumulates and impairs the prediction accuracy, especially for long-range prediction. 
Also, when applying most discrete GNNs to learn over multiple systems under different dynamical laws (environments), they usually retrain the GNNs individually for dealing with each specific system environment~\cite{jure,NRI}, which yields a large computational cost.

\subsection{Ordinary Differential Equations (ODEs) for
Multi-agent Dynamical Systems}
The dynamic nature of a multi-agent system can be captured by a series of nonlinear first-order ordinary differential equations (ODEs), which describe the co-evolution of states for a set of $N$ dependent variables (agents) over continuous time $t\in\mathbb{R}$ as ~\cite{fluid,latentODE}: $\dot{\bm{z}}_{i}^{t}:=\frac{d \bm{z}_{i}^{t}}{d t}=g\left(\bm{z}_{1}^{t}, \bm{z}_{2}^{t} \cdots \bm{z}_{N}^{t}\right)$. Here $\bm{z}_i^t\in\mathbb{R}^d$ denotes the state variable for agent $i$ at timestamp $t$ and $g$ denotes the ODE function that drives the system move forward. Given the initial states $\bm{z}_{1}^{0}, 
\cdots \bm{z}_{N}^{0}$ for all agents and the ODE function $g$, any black box numerical ODE solver such as Runge-Kuttais~\cite{solver} can solve the ODE initial-value problem (IVP), of which the solution $\bm{z}_i^T$ can be evaluated at any desired time as shown in Eqn~\ref{eq:ode}. 
\begin{equation}
    \bm{z}_{i}^{T}=\bm{z}_{i}^{0}+\int_{t=0}^{T} g\left(\bm{z}_{1}^{t}, \bm{z}_{2}^{t} \cdots \bm{z}_{N}^{t}\right) d t
    \label{eq:ode}
\end{equation}
Traditionally, the ODE function $g$ is usually hand-crafted based on some domain knowledge such as in robot motion control~\cite{robots_ode} and fluid dynamics ~\cite{fluid_ode}, which is hard to specify without knowing too much about the underlying principles. Even if the exact ODE functions are given, they are usually hard to scale as they require complicated numerical integration~\cite{mokalled2016finite,jure}. Some recent studies~\cite{latentODE,LG-ODE,CG-ODE} propose to parameterize it with a neural network and learn it in a data-driven way. They combine the expressive power of neural networks along with the principled modeling of ODEs for dynamical systems, which have achieved  promising results in various applications~\cite{LG-ODE,CG-ODE,latentODE}.

\subsection{GraphODE for Dynamical Systems}
To model the complex interplay among agents in a dynamical system, researchers have recently proposed to combine ODE with GNNs, which has been shown to achieve superior performance in long-range predictions~\cite{LG-ODE,CG-ODE,ndcn}. In~\cite{ndcn}, an encoder-processor-decoder architecture is proposed, where an encoder first computes the latent initial states for all agents individually based on their first observations. Then an ODE function parameterized by a GNN predicts the latent trajectories starting from the learned initial states. Finally, a decoder extracts the predicted dynamic features based on a decoding function that takes the predicted latent states as input. Later on, a  Graph-ODE framework has been proposed~\cite{LG-ODE,CG-ODE} which follows the structure of variational autoencoder~\cite{VAE}. They assume an approximated posterior distribution over the latent initial state for each agent, which is learned based on the whole historical trajectories instead of a single point as in~\cite{ndcn}. The encoder computes the approximated posterior distributions for all agents simultaneously considering their mutual influence and then sample the initial states from them. Compared with~\cite{ndcn}, they are able to achieve better prediction performance, especially in the long range, and are also capable of handling the dynamic evolution of graph structures~\cite{CG-ODE} which is assumed to be static in~\cite{ndcn}.

We follow a similar framework to this line but aim at generalizing GraphODE to model multiple systems across environments. 

%% file: 04-Method.tex
\section{Method}
\begin{figure*}[ht]
    \centering
  \includegraphics[width=1\linewidth]{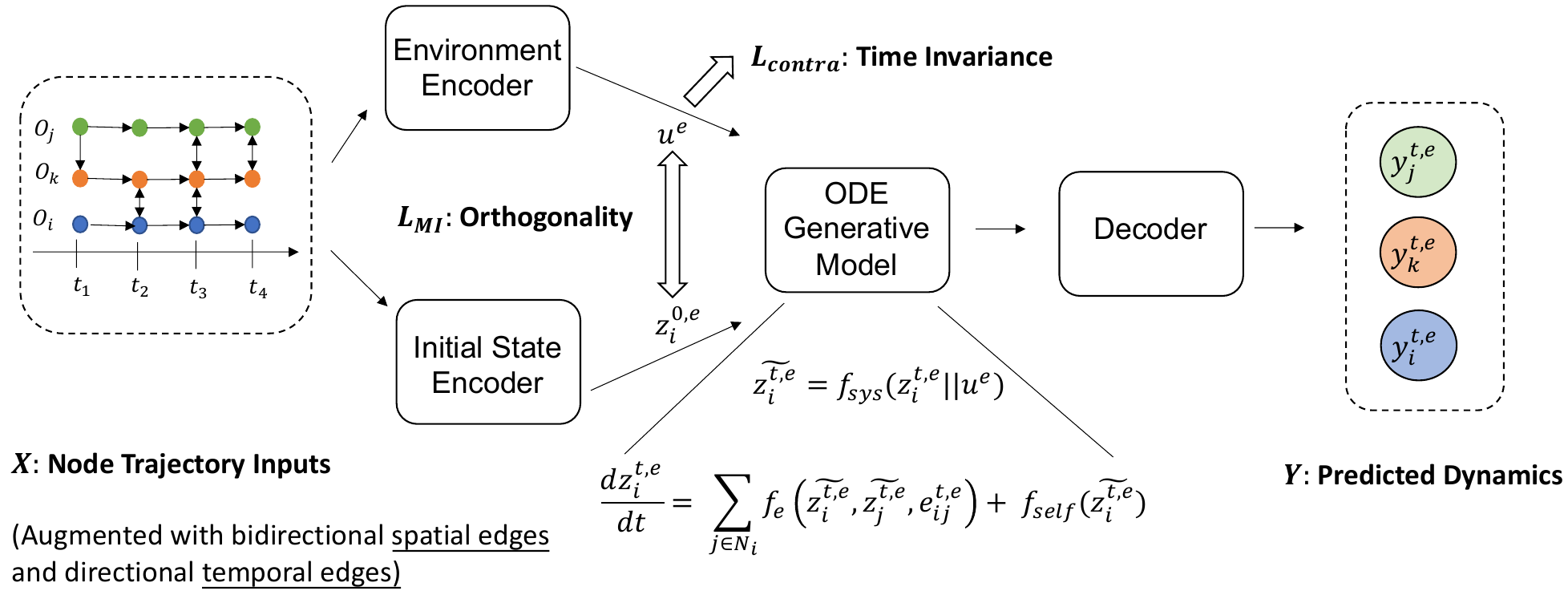}
  \caption{The overall framework of \model~~ consists of four modules. First, an initial state encoder computes the latent initial states for all agents simultaneously by constructing a temporal graph from the input trajectories. Additionally, an environment encoder computes the latent representations for exogenous factors that are distinct for each environment. Then, the generative model defined by a GNN-based ODE function calls the solver to output the predicted latent states for agents in the future, where the learned exogenous factors are incorporated into the ODE function. Finally, a decoder generates the predicted dynamics for each agent based on the decoding likelihood determined by the latent states. Two regularization terms are added to preserve the orthogonality of two encoders and the time-invariant property of the environment encoder.}
  \label{fig:framework}
 \end{figure*}
 
In this section, we present Generalized Graph ODE (\model~) for learning complex system dynamics across environments. As depicted in Figure~\ref{fig:framework}, \model~~ consists of four main components that are trained jointly: (1) an initial state encoder for inferring the latent initial states for all agents simultaneously; (2) an environment encoder which learns the latent representations for exogenous factors; (3) a generative model defined by a GNN-based ODE function that is shared across environments for modeling the continuous interaction among agents in the latent space. The distinct latent exogenous factors learned
for each environment are incorporated into the ODE function to
account for their discrepancies, and (4) a decoder that extracts the predicted dynamic features based on a decoding function. We now introduce each component in detail.

\subsection{Initial State Encoder}\label{sec:initial}
Given the observed trajectories $X^{t_{1:K},e}$, the initial state encoder computes a posterior distribution of latent initial state $q_{\phi}\left(\boldsymbol{z}_{i}^{0,e} \mid X^{t_{1:K},e}\right)$ for each agent, from which $\bm{z}_i^{0,e}$ is sampled. The latent initial state $\bm{z}_i^{0,e}$ for each agent determines the starting point for the predicted trajectory. We assume the prior distribution $p(\bm{z}_i^{0,e})$ is a standard normal distribution, and use Kullback–Leibler divergence term in the loss function to add significant regularization towards how the learned distributions look like, which differs VAE from other autoencoder frameworks~\cite{NRI,CG-ODE,latentODE}.
In multi-agent dynamical systems, agents are highly-coupled and influence each other. Instead of learning such distribution separately for each agent, such as using an RNN~\cite{latentODE} to encode the temporal pattern for each individual trajectory, we compute the posterior distributions for all agents simultaneously (similar to~\cite{CG-ODE}). 
Specifically, we fuse all trajectories as a whole into a temporal graph to consider both the temporal patterns of individual agents and the  mutual interaction among them, where each node is an observation of an agent at a specific timestamp. Two types of edges are constructed, which are (1) spatial edges $\mathcal{V}^t$ that are among observations of interacting agents at each timestamp if the Euclidean distance between the agents' positions $r_{ij}^{t,e} = ||\bm{p}_i^{\te} - \bm{p}_j^{\te}||_2$ is within a (small) connectivity radius $R$; and (2) temporal edges that preserve the autoregressive nature of each trajectory, defined between two consecutive observations of the same agent. Note that spatial edges are bidirectional while temporal edges are directional to preserve the autoregressive nature of each trajectory, as shown in Figure~\ref{fig:framework}. Based on the constructed temporal graph, we learn the latent initial states for all agents through a two-step procedure: (1) dynamic node representation learning that learns the  representation $\bm{h}_{i}^{\te}$ for each observation node whose feature vector is $\bm{x}_i^{\te}$. (2) sequence representation learning that summarizes each observation sequence (trajectory) into
a fixed-dimensional vector through a self-attention mechanism.

\subsubsection{Dynamic Node Representation Learning.} We first conduct dynamic node representation learning on the temporal graph through an attention-based spatial-temporal GNN defined as follows:
\begin{equation*}
    \bm{h}_j^{l+1(t,e)}=\bm{h}_j^{l(t,e)}+\sigma\left(\sum_{i^{(t^\prime,e)} \in \mathcal{N}_{j}^{(\te)}}\alpha_{i}^{l\left(t',e\right) \rightarrow j\left(t,e\right)}\times \bm{W}_v\hat{\bm{h}}_{i}^{l\left(t',e\right)}\right)
\end{equation*}
\begin{equation}
    \alpha_{i}^{l\left(t',e\right) \rightarrow j\left(t,e\right)} = \left(\bm{W}_{k}\hat{\bm{h}}_{i}^{l\left(t',e\right)}\right)^{T}\left(\bm{W}_{q} \bm{h}_j^{l(t,e)}\right) \cdot \frac{1}{\sqrt{d}}
    \label{eq:encoder_GNN}
\end{equation}
\begin{equation*}
    \hat{\bm{h}}_{i}^{l\left(t',e\right)}= \bm{h}_{i}^{l\left(t',e\right)} + \text{TE}(t'-t)
\end{equation*}
\begin{equation*}
    \operatorname{TE}(\Delta t)_{2i}=\sin \left(\frac{\Delta t} {10000^{2 i / d}}\right),~\operatorname{TE}(\Delta t)_{2i+1}=\cos \left(\frac{\Delta t} {10000^{2 i / d}}\right)
\end{equation*}
where $\sigma(\cdot)$ is a non-linear activation function; $d$ is the dimension of node embeddings. The node representation is computed as a weighted summation over its neighbors plus residual connection where the attention score is a transformer-based~\cite{attention} dot-product of node representations by the use of value, key, query projection matrices $\bm{W}_v,\bm{W}_k,\bm{W}_q$. The learned attention scores are normalized via softmax across all neighbors. Here $\bm{h}_j^{l(t,e)}$ is the representation of agent $j$ at time $t$ in the $l$-th layer. $\bm{h}_i^{l(t^\prime,e)}$ is the general representation for a neighbor which is connected either by a temporal edge (where $t'<t$ and $i=j$) or a spatial edge (where $t=t'$ and $i\neq j$) to the observation $\bm{h}_j^{l(t,e)}$. We add temporal encoding ~\cite{attention,latentODE} to each neighborhood node representation in order to distinguish the message delivered via spatial and temporal edges respectively. Finally, we stack $L$ layers to get the final representation for each observation node as : $\bm{h}_{i}^{\te} = \bm{h}_{i}^{L\left(t,e\right)}$. 

\subsubsection{Sequence Representation Learning} We then employ a self-attention mechanism to generate the sequence representation $\bm{m}_i^e$ for each agent, which is used to compute the mean $\bm{\mu_i^{0,e}}$ and variance $\bm{\sigma_i^{0,e}}$ of the approximated posterior distribution of the agent's initial state. Compared with recurrent models such as RNN, LSTM~\cite{LSTM}, it offers better parallelization for accelerating training speed and in the meanwhile alleviates the vanishing/exploding gradient problem brought by long sequences~\cite{Dygnn}.

We follow~\cite{CG-ODE} and compute the sequence representation $\bm{m}_i^e$ as a weighted sum of observations for agent $i$:
\begin{equation}
\begin{aligned}
    \bm{m}_{i}^e=\frac{1}{K} \sum_{t} \sigma\left((\bm{a}_{i}^{e})^T \hat{\bm{h}}_{i}^{\te}\hat{\bm{h}}_{i}^{\te}\right),~\bm{a}_{i}^e=\tanh \left(\left(\frac{1}{K} \sum_{t} \hat{\bm{h}}_{i}^{\te}\right) \bm{W}_{a}\right),\\
\end{aligned}
\label{eq:encoder_sequence}
\end{equation}
 where $\bm{a}_i^e$ is the average of observation representations with a nonlinear transformation $\bm{W}_a$ and $\hat{\bm{h}}_{i}^{\te} = \bm{h}_{i}^{\te} + \text{TE}(t)$. $K$ is the number of observations for each trajectory. Then the initial state is drawn from the
approximated posterior distribution as:

\begin{equation}
    \label{eq:sample}
    \begin{aligned}
            q_{\phi}\left(\boldsymbol{z}_{i}^{0,e} \mid X^{t_{1:K},e}\right)=\mathcal{N}\left(\bm{\mu}_{i}^{0,e}, \boldsymbol{\sigma}_{i}^{0,e}\right)&,~\bm{\mu}_{i}^{0,e}, \boldsymbol{\sigma}_{i}^{0,e}=f_{\text{trans}}\left(\bm{m}_{i}^{e}\right)\\
            \bm{z}_{i}^{0,e} \sim p\left(\bm{z}_{i}^{0,e}\right)&\approx q_{\phi}\left(\boldsymbol{z}_{i}^{0,e} \mid X^{t_{1:K},e}\right)
    \end{aligned}
\end{equation}
where $f_{\text{trans}}$ is a simple Multilayer Perceptron (MLP) whose output vector is equally split into two halves to represent the mean and variance respectively.

\subsection{Environment Encoder}~\label{sec:env_encoder}
The dynamic nature of a multi-agent system can be largely affected by some exogenous factors from its environment such as gravity, temperature, etc. These exogenous factors can span over a wide range of settings and are sometimes latent and not observable. To make our model generalize across environments, we design an environment encoder to learn the effect of the exogenous factors automatically from data to account for the discrepancies across environments. Specifically, we use the environment encoder to learn the representations of exogenous factors from observed trajectories and then incorporate the learned vector into the ODE function which is shared across environments and defines how the system evolves over time. In this way, we use a shared ODE function framework to capture the commonalities across environments while preserving the differences among them with the environment-specific latent representation, to improve model generalization performance. It also allows us to learn the exogenous factors of an unseen environment based on only its leading observations. 
We now introduce the environment encoder in detail.

The exogenous factors would pose influence on all agents within a system. On the one hand, they will influence the self-evolution of each individual agent. For example, temperatures would affect the velocities of agents. On the other hand, they will influence the pair-wise interaction among agents. For example, temperatures would also change the energy when two particles collide with each other. The environment encoder $f_{\text{enc}}^{\text{env}}$ therefore learns the latent representation of exogenous factors $\bm{u}^e$ by jointly consider the trajectories from all agents, i.e. $f_{\text{enc}}^{\text{env}}: X^{t_{1:K},e}\rightarrow \bm{u}^e$. Specifically, we learn an environment-specific latent vector from the aforementioned temporal graph in Sec~\ref{sec:initial} that is constructed from observed trajectories. The temporal graph contains both the information for each individual trajectory and the mutual interaction among agents through temporal and spatial edges. 
To summarize the whole temporal graph into a vector $\bm{u}^e$, we attend over the sequence representation $\bm{m}_i^e$ for each trajectory introduced in Sec~\ref{sec:initial} as:
\begin{equation}
\begin{aligned}
    \bm{u}^e=\frac{1}{N} \sum_{i} \sigma\left((\bm{b}^{e})^T \bm{m}_{i}^{e}\bm{m}_{i}^{e}\right),~\bm{b}^e=\tanh \left(\left(\frac{1}{N} \sum_{i} \bm{m}_i^e\right) \bm{W}_{b}\right),\\ 
\end{aligned}
\label{eqn:environment}
\end{equation}
where $\bm{W}_{b}$ is a transformation matrix and the attention weight is computed based on the average sequence representation with nonlinear transformation similar as in Eqn~\eqref{eq:encoder_sequence}. Note that we use different parameters to compute the sequence representation $\bm{m}_i^e$ as opposed to the initial state encoder. The reason is that the semantic meanings of the two sequence representations are different: one is for the latent initial states and another is for the exogenous factors.

\subsubsection{Time Invariance.}
A desired property of the learned representation for exogenous factors $\bm{u}^e$ is that it should be time-invariant towards the input trajectory time window. In other words, for the same environment, if we chunk the whole trajectories into several pieces, the inferred representations should be similar to each other as they are describing the same environment.

\begin{figure}[ht]
    \centering
  \includegraphics[width=1\linewidth]{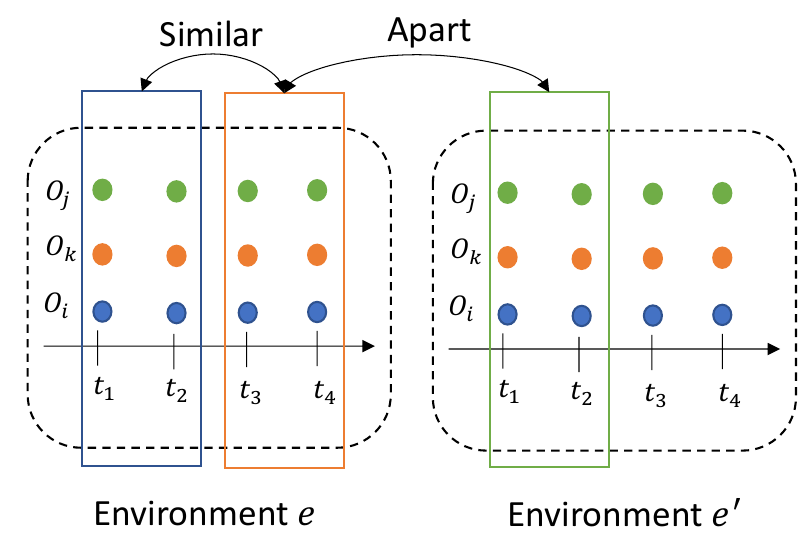}
  \caption{Temporal properties of the environment encoder. We use contrastive learning loss to force the latent exogenous factors learned from different windows within the same environment to be close to each other, and from different environments to be apart from each other.}
  \label{fig:temporal}
 \end{figure}

To achieve this, we design a contrastive learning loss to guide the learning process of the exogenous factors. As shown in Figure~\ref{fig:temporal}, we force the learned exogenous factor representations to be similar if they are generated based on the trajectories from the same environment (positive pairs), and to be apart from each other if they are from different environments (negative pairs). Specifically, we define the contrastive leanring loss as follows:

\begin{equation}
\mathcal{L}_{\text {contra }}=-\log \frac{\exp \left(\operatorname{sim}\left(f_{\text {enc }}^{\text {env }}\left(X^{t_1: t_2, e}\right), f_{\text {enc }}^{\text{env }}\left(X^{t_3: t_4, e}\right)\right) / \tau\right)}{\sum_{e^{\prime} \neq e} \exp \left(\operatorname { s i m } \left(f_{\text {enc }}^{\text {env }}\left(X^{t_1: t_2, e}, f_{\text {enc }}^{\text {env }}\left(X^{t_5: t_6, e^{\prime}}\right) / \tau\right)\right.\right.}
\label{eqn:contrast}
\end{equation}
where $\tau$ is a temperature scalar and sim$(\cdot, \cdot)$ is cosine similarity between two vectors. Note that the lengths of the observation sequences can vary. The detailed generation process for positive and negative pairs can be found in Appendix~\ref{sec:contra_sampling}.

\subsubsection{Orthogonality.}
\model~~ features two encoders that take the input of observed trajectories $X^{t_{1:K},e}$ for learning the latent initial states and the latent exogenous factors respectively. As they are designed for different purposes but are both learned from the same input, we disentangle the learned representations from them via a regularization loss defined via mutual information minimization.

Mutual information measures the dependency between two random variables $X,Z$~\cite{zhu2021learning}. Since we are not interested in the exact value of the mutual information, a lower bound derived from Jensen Shannon Divergence~\cite{hjelm2018learning} could be formulated as
\begin{equation}
I_{\mathrm{JSD}}(X, Z)= E_{P_{X Z}}[-\operatorname{sp}(-M(x, z))] -E_{P_X P_Z}[\operatorname{sp}(M(x, z))],
\end{equation}
where $P_X P_Z$ is the product of the marginal distributions and $P_{X Z}$ is the joint distribution. $sp(w)=log(1+e^w)$ and $M$ is a discriminator modeled by a neural network to compute the score for measuring their mutual information.

According to recent literature~\cite{zhu2021learning,hjelm2018learning,sanchez2020learning}, the sample pair (positive pairs) $(x,z)$ drawn from the joint distribution $P_{X Z}$ are different representations of the same data sample, and the sample pair (negative pairs) drawn from $P_X P_Z$ are different representations from different data samples. We therefore attempt to minimize the mutual information from the two encoders as follows
\begin{equation}
     \mathcal{L}_{\text{MI}}
     =\mathbb{E}_{e\in E,i}[-sp(-\Psi(\bm{z}_i^{0,e}, \bm{u}^e))]-\mathbb{E}_{e\in E\times e'\in E,i}[sp(\Psi(\bm{z}_i^{0,e}, \bm{u}^{e'}))]   
  \label{eqn:mi}
\end{equation}
 where $\Psi$ is a MLP-based discriminator. Specifically, we force the latent initial states $\bm{z}_i^{0,e}$ for all agents from environment $e$ to be dissimilar to the learned exogenous factors $\bm{u}^e$. And construct negative pairs by replacing the learned exogenous factors from another environment as $\bm{u}^{e'}$. The generation process for positive and negative pairs can be found in Appendix~\ref{sec:MI_sampling}.

\subsection{ODE Generative Model and Decoder}
\subsubsection{ODE Generative Model}
After describing the initial state encoder and the environment encoder, we now define the ODE function that drives the system to move forward. The future trajectory of each agent can be determined by two important factors: the potential influence received from its neighbors in the interaction graph and the self-evolution of each agent. For example, in the n-body system, the position of each agent can be affected both by the force from its connected neighbors and its current velocity which can be inferred from its historical trajectories. 
Therefore, our ODE function consists of two parts: a GNN that captures the continuous interaction among agents and the self-evolution of the node itself. One issue here is how can we decide the neighbors for each agent in the ODE function as the interaction graph is evolving, the neighbors for each agent are dynamically changing based on their current positions, which are implicitly encoded in their latent state representations $\bm{z}_i^{t,e}, \bm{z}_j^{t,e}$. We propose to first decode the latent node representations $\bm{z}_i^{t,e}, \bm{z}_j^{t,e}$ with a decoding function $f_{\text{dec}}$ to obtain their predicted positions $\bm{p}_i^{t,e}, \bm{p}_j^{t,e}$ at current timestamp. Then we determine their connectivity based on whether their Euclidean distance $r_{ij}^{t,e} = ||\bm{p}_i^{\te} - p_j^{t,e}||_2$ is within the predefined radius $R$. This can be computed efficiently by using a multi-dimensional index structure such as the $k\text{-}d$ tree. The decoding function $f_{\text{dec}}$ is the same one that we will use in the decoder.  

To incorporate the influence of exogenous factors, we further incorporate $\bm{u}^e$ into the general ODE function to improve model generalization ability as:

\begin{equation}
    \begin{aligned}
          \frac{d \bm{z}_i^{\te}}{dt} = g\left(\bm{z}_{1}^{\te}, \bm{z}_{2}^{\te} \cdots \bm{z}_{N}^{\te}\right) &= \sum_{j\in \mathcal{N}_i} f_{\text{GNN}}(\widetilde{\bm{z}}_i^{\te}, \widetilde{\bm{z}}_j^{\te}) + f_{\text{self}}(\widetilde{\bm{z}}_i^{\te})\\
         \widetilde{\bm{z}}_i^{\te} &= f_{\text{env}}(\bm{z}_i^{\te}||\bm{u}^e)
    \end{aligned}
    \label{eqn:ode}
\end{equation}
where $||$ denotes concatenation and $f_{\text{GNN}}$ can be any GNN that conducts message passing among agents. $f_{\text{self}}, f_{\text{env}}$ are implemented as two MLPs respectively.  In this way, we learn the effect of latent exogenous factors from data without supervision where the latent representation $\bm{u}^e$ is trained end-to-end by optimizing the prediction loss.

\subsubsection{Decoder}
Given the ODE function $g$ and agents' initial states $\bm{z}_i^{0,e}$ for $i=1,2\cdots N$, the latent trajectories for all agents are determined, which can be solved via any black-box ODE solver. Finally, a decoder generates the predicted dynamic features based on the decoding probability $p(\bm{y}_{i}^{t,e} | \bm{z}_{i}^{t,e})$ computed from the decoding function $f_{\text{dec}}$ as shown in Eqn~\ref{eq:generative}. We implement $f_{\text{dec}}$ as a simple two-layer MLP with nonlinear activation. It outputs the mean of the normal distribution $ p(\bm{y}_{i}^{t,e} | \bm{z}_{i}^{t,e})$, which we treat as the predicted value for each agent.
 \begin{equation}
\begin{aligned}
    \bm{z}_{i}^{t_1,e}\cdots \bm{z}_{i}^{t_T,e} &= \text{ODESolve}(g, [\bm{z}_1^{0,e},\bm{z}_2^{0,e}\cdots \bm{z}_N^{0,e}],(t_1\cdots t_T))\\
     \bm{y}_i^{t,e} &\sim p(\bm{y}_{i}^{t,e} | \bm{z}_{i}^{t,e}) = f_{\text{dec}}(\bm{z}_{i}^{t,e})
\end{aligned}
    \label{eq:generative}
\end{equation}

\subsection{Training}
We now introduce the overall training procedure of \model~~. For each training sample, we split it into two halves along the time, where we condition on the first half $[t_1,t_K]$ in order to predict dynamics in the second half $[t_{K+1},t_T]$. Given the observed trajectories $X^{t_{1:K},e}$, we first run the initial state encoder to compute the latent initial state $\bm{z}_i^{0,e}$ for each agent, which is sampled from the approximated posterior distribution $q_{\phi}\left(\boldsymbol{z}_{i}^{0,e} \mid X^{t_{1:K},e}\right)$. We then generate the latent representations of exogenous factors $\bm{u}^e$ from the environment $e$ via the environment encoder. Next, we run the ODE generative model that incorporates the latent exogenous factors to compute the latent states for all agents in the future. Finally, the decoder outputs the predicted dynamics for each agent.

We jointly train the encoders, ODE generative model, and decoder in an end-to-end manner. The loss function consists of three parts: (1) the evidence lower bound (ELBO) which is the addition of the reconstruction loss for node trajectories and the KL divergence term for adding regularization to the inferred latent initial states for all agents. We use $\bm{Z}^{0,e}$ to denote the latent initial state matrix of all N agents. The standard VAE framework is trained to maximize ELBO so we take the negative as the ELBO loss; (2) the contrastive learning loss for preserving the time invariance properties of the learned exogenous factors; (3) the mutual information loss that disentangles the learned representations from the two encoders. $\lambda_1, \lambda_2$ are two hyperparameters for balancing the three terms. We summarize the whole procedure in Appendix~\ref{appendix:algo}.

\begin{equation}
    \mathcal{L} = \mathcal{L}_{\text{ELBO}} + \lambda_1\mathcal{L}_{\text{contra}} + \lambda_2 \mathcal{L}_{MI}
    \label{eqn:loss_all}
\end{equation}

\begin{equation}
\begin{aligned}
      \mathcal{L}_{\text{ELBO}(\theta,\phi)} = -\mathbb{E}_{\bm{Z}^{0,e} \sim \prod_{i=1}^{N}q_{\phi}\left(\boldsymbol{z}_{i}^{0,e} \mid X^{t_{1:K},e}\right)}[\log p_{\theta}(Y^{t_{K+1:T},e})]\\
    +\mathrm{KL}[\prod_{i=1}^{N}q_{\phi}(\bm{z}_i^{0,e} |X^{t_{1:K},e})\| p(\bm{Z}^{0,e})]
\end{aligned}
\end{equation}

%% file: 05-Experiment.tex
\section{Experiments}

\begin{table*}[ht]
\caption{Mean Square Error (MSE) of rollout trajectories with varying prediction lengths. The transductive setting evaluates the testing sequences whose environments are seen during training. The inductive setting evaluates new systems with unseen environments during training. The best results are bold-faced.}
\begin{tabular}{l|ccc|ccc|ccc|ccc}
\toprule
\hline
\multicolumn{1}{c|}{Dataset}            & \multicolumn{3}{c|}{\begin{tabular}[c]{@{}c@{}}Lennard-Jones potential\\ Transductive MSE ($10^{-2}$)\end{tabular}} & \multicolumn{3}{c|}{\begin{tabular}[c]{@{}c@{}}Lennard-Jones potential\\ Inductive MSE ($10^{-1}$)\end{tabular}} & \multicolumn{3}{c|}{\begin{tabular}[c]{@{}c@{}}Water\\ Transductive MSE ($10^{-3}$)\end{tabular}} & \multicolumn{3}{c}{\begin{tabular}[c]{@{}c@{}}Water\\ Inductive MSE ($10^{-2}$)\end{tabular}} \\
\multicolumn{1}{c|}{Rollout Percentage} & 30\%                               & 60\%                                & 100\%                               & 30\%                               & 60\%                              & 100\%                              & 30\%                           & 60\%                           & 100\%                          & 30\%                         & 60\%                        & 100\%                       \\ \hline
LSTM                                    & 6.73                               & 20.69                               & 31.88                               & 1.64                               & 8.82                              & 18.01                              & 4.87                           & 23.09                          & 30.44                          & 1.01                         & 6.72                        & 14.79                       \\
NRI                                     & 5.83                               & 17.99                               & 28.18                               & 1.33                               & 4.34                              & 13.97                              & 3.87                           & 19.64                          & 26.34                          & 0.83                         & 3.84                        & 10.59                       \\
NDCN                                    & 5.99                               & 17.54                               & 27.06                               & 1.35                               & 4.27                              & 12.37                              & 3.95                           & 18.76                          & 24.33                          & 0.85                         & 3.79                        & 10.11                       \\
CG-ODE                                     & 5.43                      & 17.01                               & 26.01                               & 1.32                               & 4.25                              & 12.03                              & 3.41                  & 18.13                          & 23.62                          & 0.80                         & 3.64                        & 9.91                        \\
SocialODE                                     & 5.62                      & 17.23                               & 26.89                               & 1.34                               & 4.26                              & 12.44                              & 3.68                  & 18.42                          & 23.77                          & 0.84                         & 3.70                        & 10.01                        \\
GNS                                     & \textbf{5.03}                      & 16.28                               & 25.44                               & 1.28                               & 4.23                              & 11.88                              & \textbf{3.17}                  & 17.88                          & 23.14                          & 0.76                         & 3.45                        & 9.78                        \\
\model~                                  & 5.18                               & \textbf{16.03}                      & \textbf{24.97}                      & \textbf{1.10}                      & \textbf{3.98}                     & \textbf{10.77}                     & 3.20                           & \textbf{16.94}                 & \textbf{22.58}                 & \textbf{0.63}                & \textbf{3.11}               & \textbf{8.02}               \\ \hline
-w/o $\mathcal{L}_\text{contra}$                          & 5.32                               & 17.03                               & 26.53                               & 1.30                               & 4.25                              & 12.13                              & 3.32                           & 18.03                          & 23.01                          & 0.75                         & 3.58                        & 10.03                       \\
-w/o$\mathcal{L}_{MI}$                               & 5.45                               & 17.25                               & 26.11                               & 1.32                               & 4.11                              & 11.76                              & 3.43                           & 18.32                          & 22.95                          & 0.78                         & 3.51                        & 9.88                        \\
shared encoders                         & 5.66                               & 17.44                               & 26.79                               & 1.33                               & 4.46                              & 12.22                              & 3.55                           & 18.57                          & 23.55                          & 0.81                         & 3.66                        & 10.08                       \\ \hline
\bottomrule
\end{tabular}
\label{table:main_results}
\end{table*}

\subsection{Experiment Setup}
\subsubsection{Datasets} 
We illustrate the performance of our model across two physical simulations that exhibit different system dynamics over time: (1) The Water dataset~\cite{jure}, which describes the fluid dynamics of water within a container. Containers can have different shapes and numbers of ramps with random positions inside them, which we view as different environments. The dataset is simulated using the material point method (MPM), which is suitable for simulating the behavior of interacting, deformable materials such as solids, liquids, gases~\footnote{\url{https://en.wikipedia.org/wiki/Material_point_method}}. For each data sample, the number of particles can vary but the trajectory lengths are kept the same as 600. The input node features are 2-D positions of particles, and we calculate the velocities and accelerations as additional node features using finite differences of these positions. The total number of data samples (trajectories) is 1200 and the number of environments is 68, where each environment can have multiple data samples with different particle initializations such as positions, velocities, and accelerations. 
(2) The Lennard-Jones potential dataset~\cite{jones1924determination}, which describes the soft repulsive and attractive interactions between simple atoms and molecules~\footnote{\url{https://en.wikipedia.org/wiki/Lennard-Jones_potential}}. We generate data samples with different temperatures, which could affect the potential energy preserved within the whole system thus affecting the dynamics. We view temperatures as different environments. The total number of data samples (trajectories) is 6500 and the number of environments is 65. Under each environment, we generate 100 trajectories with different initializations. The trajectory lengths are kept the same as 100. The number of particles is 1000 for all data samples. More details about datasets can be found in Appendix~\ref{sec:appendix_data}.

\subsubsection{Task Evaluation and Data Split}
We predict trajectory rollouts across varying lengths and use Mean Square Error (MSE) as the evaluation metric.

\noindent\textbf{Task Evaluation.}
The trajectory prediction task is conducted under two settings: (1) Transductive setting, where we evaluate the test sequences whose environments are seen during training; (2) Inductive setting, where we evaluate the test sequences whose environments are not observed during training. It helps to test the model's generalization ability to brand-new systems.

\noindent\textbf{Data Split.}
We train our model in a sequence-to-sequence setting where we split the trajectory of each training sample into two parts $[t_1,t_K]$ and $[t_{K+1},t_T]$. We condition on the first part of observations to predict the second part. To conduct data split, we first randomly select 20$\%$ environments whose trajectories are all used to construct the testing set $X_{\text{test}}^{\text{Induct}}$ in the inductive setting. For the remaining trajectories that cover the 80$\%$ environments, we randomly split them into three partitions: $80\%$ for the training set $X_{\text{train}}$, $10\%$ for the validation set $X_{\text{val}}$ and $10\%$ for the testing set in the transductive setting $X_{\text{test}}^{\text{trans}}$. In other words, we have two test sets for the inductive and transductive settings respectively,  one training set and one validation set. 
To fully utilize the data points within each trajectory, we generate training and validation samples by splitting each trajectory into several chunks that can overlap with each other, using a sliding window. The sliding window has three hyperparameters: the observation length and prediction length for each sample, and the interval between two consecutive chunks (samples). Specifically, for the Water dataset, we set the observation length as 50 and the prediction length as 150. We obtain samples from each trajectory by using a sliding window of size 200 and setting the sliding interval as 50. For the Lennard-Jones potential dataset, we set the observation length as 20, the prediction length as 50, and the interval as 10. The procedure is summarized in Appendix~\ref{sec:appendix_split}. During evaluations for both settings, we ask the model to roll out over the whole trajectories without further splitting, whose prediction lengths are larger than the ones during training. The observation lengths during testing are set as 20 for the Lennard-Jones potential dataset and 50 for the Water dataset across the two settings. 

\begin{figure*}[ht]
    \centering
  \includegraphics[width=1\linewidth]{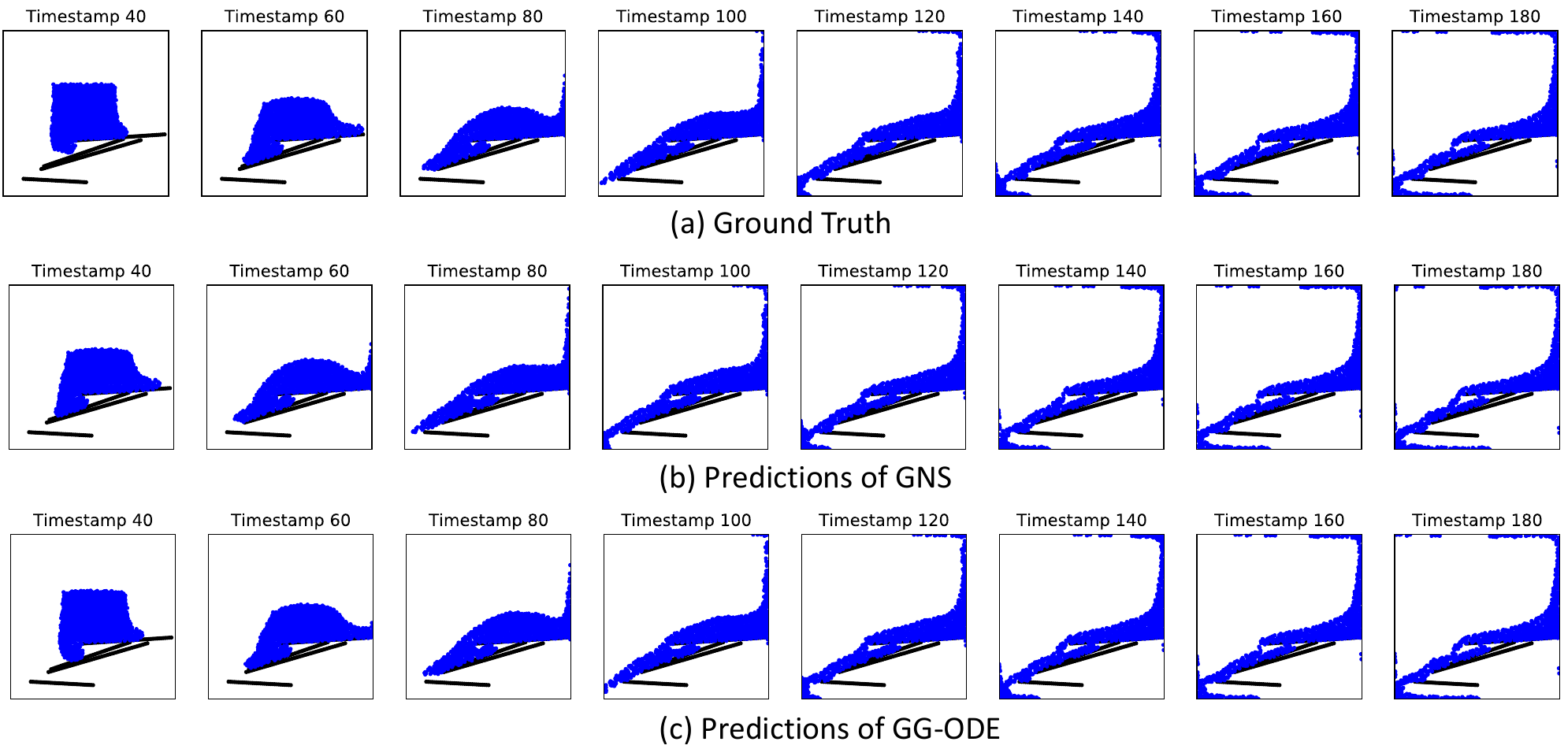}
  \caption{Visualization of the transductive prediction results for the Water dataset. Black lines are ramps within the container. The length of the observation sequence is set as 20. GNS makes less accurate predictions compared with GG-ODE.}
  \label{fig:Visualization}
\end{figure*}

\subsection{Baselines}
We compare both discrete neural models as well as continuous neural models where they do not have special treatment for modeling the influence from different environments. For discrete ones we choose: NRI~\cite{NRI} which is a discrete GNN model that uses VAE to infer the interaction type among pairs of agents and is trained via one-step predictions; GNS~\cite{jure}, a discrete GNN model that uses multiple rounds of message passing to predict every single step; LSTM~\cite{LSTM}, a classic recurrent neural network (RNN) that learns the dynamics of each agent independently. For the continuous models, we compare with NDCN~\cite{ndcn} and Social ODE~\cite{wen2022social}, two ODE-based methods that follow the encoder-processor-decoder structure with GNN as the ODE function. The initial state for each agent is drawn from a single data point instead of a leading sequence. CG-ODE~\cite{CG-ODE} which has the same architecture as our model, but with two coupled ODE functions to guide the evolution of systems.

        

\subsection{Performance Evaluation}
We evaluate the performance of our model based on Mean Square Error (MSE) as shown in Table~\ref{table:main_results}. As data samples have varying trajectory lengths, we report the MSEs over three rollout percentages regarding different prediction horizons: $30\%, 60\%, 100\%$ where $100\%$ means the model conditions on the observation sequence and predicts all the remaining timestamps. 

Firstly, we can observe that \model~ consistently outperforms all baselines across different settings when making long-range predictions, while achieving competitive results when making short-range predictions. This demonstrates the effectiveness of \model~ in learning continuous multi-agent system dynamics across environments. By comparing the performance of LSTM with other methods, we can see that modeling the latent interaction among agents can indeed improve the prediction performance compared with predicting trajectories for each agent independently. Also, we can observe the performance gap between \model~ and other baselines increase when we generate longer rollouts, showing its expressive power when making long-term predictions. This may be due to the fact that \model~ is a continuous model trained in a sequence-to-sequence paradigm whereas discrete GNN methods are only trained to make a fixed-step prediction. Another continuous model NDCN only conditions a single data point to make predictions for the whole trajectory in the future, resulting in suboptimal performance. Finally, we can see that \model~ has a larger performance gain over existing methods in the inductive setting than in the transductive setting, which shows its generalization ability to fast adapt to other unseen systems with a few data points. Figure~\ref{fig:Visualization} visualizes the prediction results under the transductive setting for the Water dataset.

\subsubsection{Ablation Studies}
To further analyze the rationality behind our model design, we conduct an ablation study by considering three model variants: (1) We remove the contrastive learning loss which forces the learned exogenous factors to satisfy the time invariance property, denoted as $-w/o \mathcal{L}_{\text{contra}}$; (2) We remove the mutual information minimization loss which reduces the variance of the learned exogenous factors from the same environment, denoted as $-w/o \mathcal{L}_{MI}$. (3) We share the parameters of the two encoders for computing the latent representation $\bm{m}_i^e$ for each observation sequence in the temporal graph, denoted as shared encoders. As shown in Table~\ref{table:main_results}, all three variants have inferior performance compared to \model~, verifying the rationality of the three key designs. Notably, when making long-range predictions, removing $\mathcal{L}_{MI}$ would cause more harm to the model than removing $\mathcal{L}_{\text{contra}}$. This can be understood as the latent initial states are more important for making short-term predictions, while the disentangled latent initial states and exogenous factors are both important for making long-range predictions.

\subsubsection{Hyperparameter Study}
\begin{figure}[ht]
    \centering
  \includegraphics[width=1\linewidth]{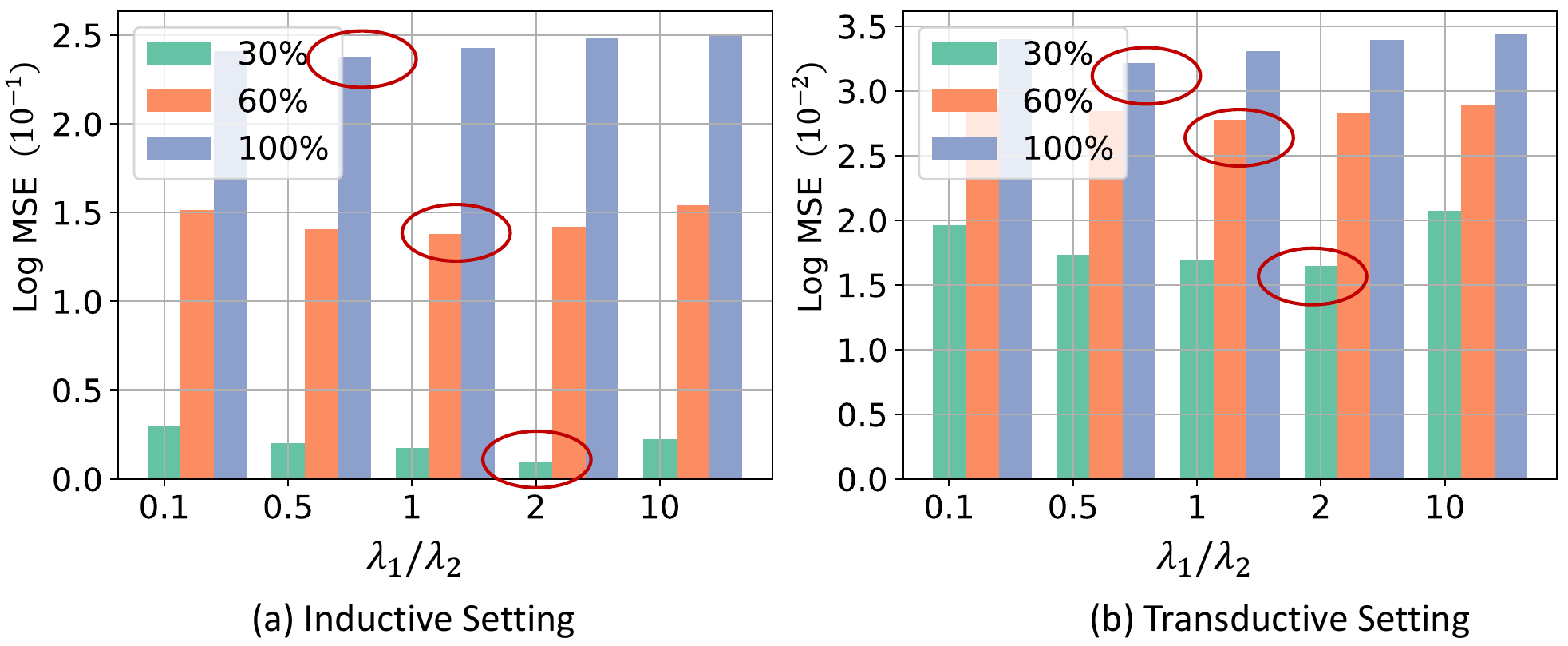}
  \caption{Effect of $\lambda_1/\lambda_2$ on the Lennard-Jones potential dataset. Best results are circled in red for each setting.}
  \label{fig:hyper}
 \end{figure}

We study the effect of $\lambda_1/\lambda_2$, which are the hyperparameters for balancing the two regularization terms that guide the learning of the two encoders, towards making predictions under different horizons. As illustrated in Figure~\ref{fig:hyper}, the optimal ratio for making $30\%, 60\%, 100\%$ rollout predictions are 2, 1,0.5 respectively, under both the transductive and inductive settings. They indicate that the exogenous factors modeling plays a more important role in facilitating long-term predictions, which is consistent with the prediction errors illustrated in Table~\ref{table:main_results} when comparing $-w/o \mathcal{L}_{MI}$ with $-w/o 
 \mathcal{L}_{\text{contra}}$. However, overly elevating $\mathcal{L}_{MI}$ would also harm the model performance, as the time invariance property achieved by $\mathcal{L}_{\text{contra}}$ is also important to guarantee the correctness of the learned latent initial states, which determines the starting point of the predicted trajectories in the future. 

\subsubsection{Sensitivity Analysis.}
\begin{figure}[ht]
    \centering
  \includegraphics[width=1\linewidth]{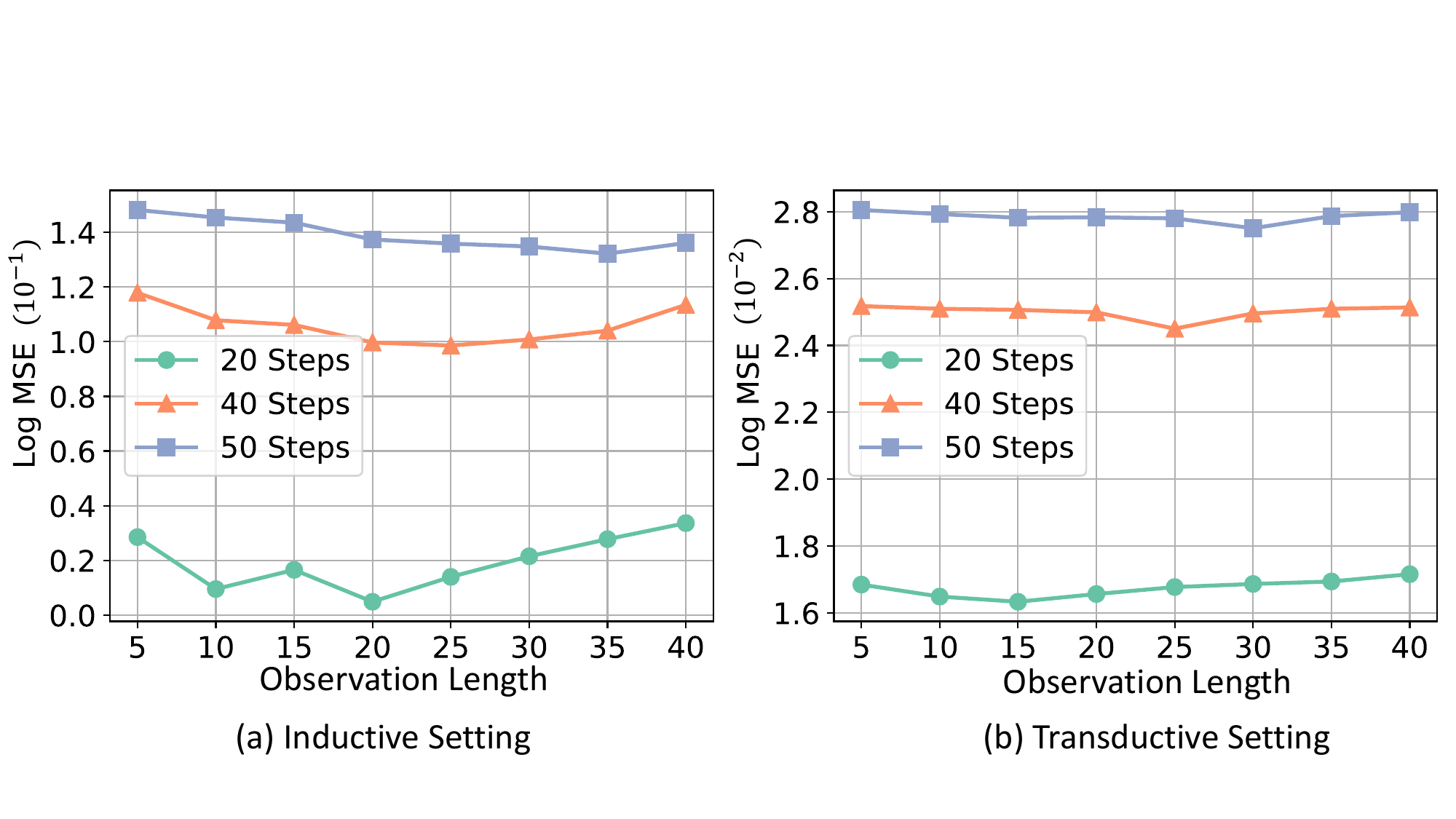}
  \caption{Effect of observation length on the Lennard-Jones potential dataset.}
  \label{fig:sensitivity}
 \end{figure}

\model~ can take arbitrary observation lengths to make trajectory predictions, as opposed to existing baselines that only condition on observations with fixed lengths. It allows the model to fully utilize all the information in the past. We then study the effect of observation lengths on making predictions in different horizons. As shown in Figure~\ref{fig:sensitivity}, the optimal observation lengths for predicting the rollouts with 20, 40, and 50 steps are 20, 25, 35 in the inductive setting, and 15, 25, 30 in the transductive setting. When predicting long-range trajectories, our model typically requires a longer observation sequence to get more accurate results. Also, for making predictions at the same lengths, the inductive setting requires a longer observation length compared with the transductive setting.

\subsection{Case Study}
\begin{figure}[ht]
    \centering
  \includegraphics[width=1\linewidth]{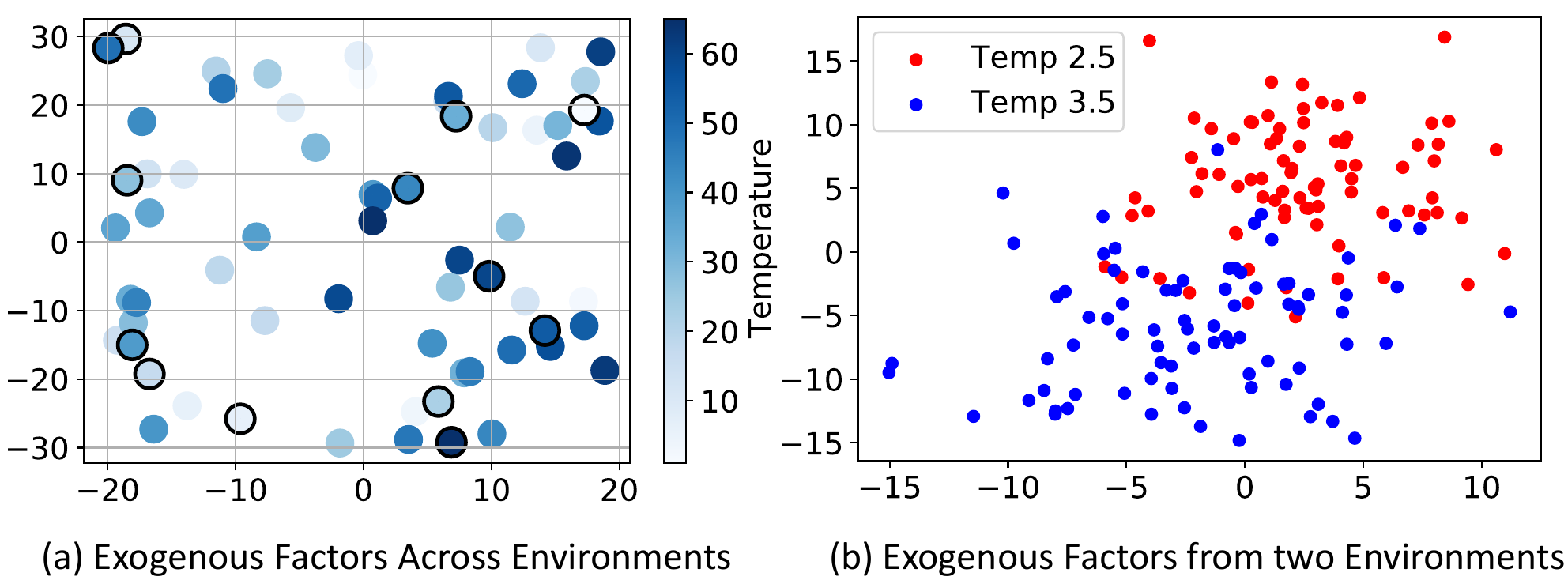}
  \caption{ T-SNE visualization of the learned exogenous factors on the Lennard-Jones potential dataset. (a) We randomly pick one data sample per temperature, where temperatures tested in the inductive setting are circled in black. (b) Visualization of  data samples from two temperatures. }
  \label{fig:case study}
 \end{figure}

We conduct a case study to examine the learned representations of the latent exogenous factors on the Lennard-Jones potential dataset. We first randomly choose one data sample for each of the 65 temperatures and visualize the learned representations of exogenous factors. As shown in Figure~\ref{fig:case study} (a), the representations of higher temperatures are closer to each other on the right half of the figure, whereas the lower temperatures are mostly distributed on the left half. Among the 65 temperatures, $20\%$ of them are not seen during training which we circled in black. We can see those unseen temperatures are also properly distributed, indicating the great generalization ability of our model. We next plot the representations for all data samples under temperatures 2.5 and 3.5 respectively as shown in Figure~\ref{fig:case study} (b). We can see that the learned representations are clustered within the two temperatures, indicating our contrastive learning loss is indeed beneficial to guide the learning process of exogenous factors.

%% file: 06-Conclusion.tex
\section{Conclusion}
In this paper, we investigate the problem of learning the dynamics of continuous interacting systems across environments. We model system dynamics in a continuous fashion through graph neural ordinary differential equations. To achieve model generalization, we learn a shared ODE function that captures the commonalities of the dynamics among environments while design an environment encoder that learns environment-specific representations for exogenous factors automatically from observed trajectories. To disentangle the representations from the initial state encoder and the environment encoder, we propose a regularization loss via mutual information minimization to guide the learning process. We additionally design a contrastive learning loss to reduce the variance of learned exogenous factors across time windows under the same environment. The proposed model is able to achieve accurate predictions for varying physical systems under different environments, especially for long-term predictions. There are some limitations though. Our current model 
only learns one static environment-specific variable to achieve model generalization. However, the environment can change over time such as temperatures. How to capture the dynamic influence of those evolving environments remain challenging.

%% file: 07-Appendix.tex
\section{Appendix}
\subsection{Datasets}~\label{sec:appendix_data}
We conduct experiments over two datasets: The Water dataset and the Lennard-Jones potential dataset. As introduced in Sec~\ref{sec:problem},  the edges between agents are assigned if the Euclidean distance between the agents' positions $r_{ij}^{t,e} = ||\bm{p}_i^{\te} - \bm{p}_j^{\te}||_2$ is within a (small) connectivity radius $R$.
The connectivity radius for the two datasets is set as 0.015 and 2.5 respectively. The number of particles is kept the same as 1000 for all trajectories in the Lennard-Jones potential dataset, while in the Water dataset, each data sample can have a varying number of particles, and the maximum number of particles is 1000.

\subsubsection{Data Split.}~\label{sec:appendix_split} Our model is trained in a sequence-to-sequence mode, where we split the trajectory of each training sample into two parts $[t_1,t_K]$ and $[t_{K+1},t_T]$. We condition on the first part of observations to predict the second part. To fully utilize the data points within each training sample, we split each trajectory into several chunks with three hyperparameters: the observation length and prediction length for each sample, and the interval between two consecutive chunks (samples). We summarize the procedure in Algorithm~\ref{al:split}, where $K$ is the number of trajectories and $d$ is the input feature dimension.

\begin{algorithm}[htbp]
\caption{Data Splitting Procedure.}
\label{al:split}
\KwIn{
Original Training trajectories  {$X_{\text{input}}\in \mathbb{R}^{K\times N\times T\times d}$};\\
Observation length $O$;
Prediction length $M$;
Interval $I$;
Trajectory length $T$.\\}
\KwOut{Training samples after splitting $X_{\text{train}}$.}
sample$\_$length = $O+M$;\\
num$\_$chunk = ($T$ - sample$\_$length )//interval + 1;\\
\For{i in range (0,K)}{
    \For{j $\text{in range}$(0,num$\_$chunk,I)}
    {Generate the split training sample as $X_{\text{input}}[i,:,j:j +
\text{sample}\_\text{length},:]$\\
Add the training sample to the training set $X_{\text{train}}$.}
    }
\end{algorithm}

\subsubsection{Input Features and Prediction Target.} For the Water dataset, the input node features are 2-D positions $p_i^{\te}$, and we additionally calculate the 2-D velocities and accelerations using finite differences of these positions as $v_i^{\te} = p_i^{\te} - p_i^{t-1,e}$, $a_i^t = v_i^{t,e} - v_i^{t-1,e} = p_i^{\te} - 2p_i^{t-1,e} + p_i^{t-2,e}$. For positions, velocities, and accelerations, we precompute their mean and variance across all samples and normalize them with z-score. For the Lennard-Jones potential dataset, the input node features are 3-D positions, velocities, and accelerations. We train the model to predict the future positions for each agent along the time for both datasets.

\subsection{Software and Experiment Environment}
We implement our model in PyTorch. All experiments are conducted on a GPU powered by an NVIDIA A100. For all datasets, we train over 100 epochs and select the one with the lowest validation loss as the reported model. We report the average results over 10 runs. Encoders, the generative model, and the decoder are jointly optimized using Adam optimizer~\cite{adam} with a learning rate 0.005. The batch size for the Water dataset is set as 128, and for the Lennard-Jones potential dataset. is set as 256.  Note that the batch size denotes the number of data samples generated as in Alg~\ref{al:split}.

\subsection{Implementation Details}
We now introduce the implementation details of our model.
\subsubsection{Initial State Encoder.}~\label{sec:initial_imple} The initial state encoder aims to infer latent initial states for all agents simultaneously via a two-step procedure: Firstly, the encoder computes the structural representation for each observation node by the use of a spatial-temporal GNN. We set the number of GNN layers $l$ as 2 and the hidden dimension as 64 across all datasets. LayerNorm~\cite{layernorm} is employed to provide training stability in our experiment. Next, a self-attention-based sequence representation learning procedure computes the sequence representation for each agent and samples the initial state from it.  We use a 2-layer MLP as $f_{\text{trans}}$ in Eqn~\ref{eq:sample} with latent dimensions as 128 and activation function as Tanh. 
\subsubsection{Environment Encoder.} The environment encoder learns the latent representations of exogenous factors based on the observed trajectories. The architecture is the same as the initial state encoder but are using two sets are parameters with the same hyperparameter settings introduced in Sec~\ref{sec:initial_imple}.

\noindent\textbf{Contrastive Learning Loss Sampling.}~\label{sec:contra_sampling}
The contrastive learning loss $\mathcal{L}_{\text{contra}}$ shown in Eqn~\ref{eqn:contrast} is designed to achieve the time invariance properties of the learned exogenous factors. Specifically, we sample the positive pairs $X^{t_1:t_2,e}, X^{t_3:t_4,e}$ using two strategies: (1) The intra-sample generation, where $^{t_1:t_2,e}, X^{t_3:t_4,e}$ are from the same training sample but representing two different time windows. We achieve this by randomly selecting two timestamps within each training sample to serve as $t_1, t_3$ respectively, and then set the window size as the observation length $L$ to get $t_2 = t_1 + L , t_4 = t_3+L$. (2) The cross-sample generation, where  $^{t_1:t_2,e}, X^{t_3:t_4,e}$ are from two different samples within the same environment $e$. Specifically, for each training sample, we first randomly choose another sample under the same environment. Then we generate $t_1, t_3$ by randomly selecting one timestamp for each of them. Finally, we calculate $t_2,t_4$ by adding the observation length. To generate negative pair $X^{t_5:t_6,e^{\prime}}$ for each $X^{t_1:t_2,e}$, we first randomly select one another environment $e'$, from which we randomly pick one data sample. Similarly, we then randomly select one timestamp within that data sample to serve as $t_5$ and then obtain $t_6$ as $t_6 = t_5 + L$. The temperature scalar $\tau$ in Eqn~\ref{eqn:contrast} is set as 0.05.

\noindent\textbf{Mutual Information Minimization Loss Sampling.}~\label{sec:MI_sampling} To disentangle the representations of the latent initial states and the exogenous factors, we design the mutual information minimization loss in Eqn~\ref{eqn:mi} as a regularization term during training. We conduct the sampling procedure for positive and negative pairs as follows: For each training sample, we pair the latent initial states $\bm{z}_i^{0,e}$ of all the $N$ agents with the learned exogenous factors $\bm{u}^e$, thus constructing $N$ positive pairs. To generate negative pairs, we randomly select another environment $e'$ and pair it with the latent initial states of all agents within one training sample. Thus we obtain the same number of positive and negative pairs during training. The  discriminator $\Psi$ is implemented as a two-layer MLP with hidden dimension and out dimension as 128 and 64 respectively.

\subsubsection{ODE Function and Solver.}
The ODE function introduced in Eqn~\ref{eq:ode} consists of two parts: the GNN $f_{\text{GNN}}$ that captures the mutual interaction among agents and $f_{\text{self}}$ that captures the self-evolution of agents. We use the following two-layer message passing GNN function as $f_{\text{GNN}}$:

\begin{equation}
\begin{aligned}
v \rightarrow e: & \mathbf{e}_{(i, j)}^{l_1(t,e)} & =f_e^1\left(\left[\mathbf{z}_i^{t,e}||\mathbf{z}_j^{\te}\right]\right) \\
e \rightarrow v: & \mathbf{z}_j^{l_1(t,e)} & =f_v^1\left(\sum_{i \neq j} \mathbf{e}_{(i, j)}^{l_1(t,e)}\right) \\
v \rightarrow e: & \mathbf{z}_j^{l_2(t,e)}  & =f_e^2\left(\left[\mathbf{z}_i^{l_1(t,e)}||\mathbf{z}_j^{l_1(t,e)} \right]\right)
\end{aligned}
\end{equation}
where $||$ denotes concatenation, $f_e^1, f_v^1, f_e^2$ are two-layer MLPs with hidden dimension size of 64. We use $\mathbf{z}_j^{l_2(t,e)}$ as output representation for agent j at timestamp t from $f_{\text{GNN}}$. The self-evolution function $f_{\text{self}}$ and the transformation function $f_{\text{env}}$ are also implemented as two-layer MLPs with hidden dimension of 64. We use the fourth-order Runge-Kutta method from torchdiffeq python package~\cite{ode} as the ODE solver, which solves the ODE systems on a time grid that is five times denser than the observed time points. We also utilize the Adjoint method described in ~\cite{ode} to reduce memory usage.

\subsection{Pseudo-Code of \model~Training}~\label{appendix:algo}

\begin{algorithm}[ht]
\caption{Generalized Graph ODE training procedure.}
\label{al:training}
\KwIn{
Observed trajectories  {$X^{t_{1:K},e}$}.}
\KwOut{Model parameters $\phi$ and $\theta$.}
\While{model not converged}{
    \For{Each training sample}{
        Separate the sequence into observed half $[T_0,T_1]$ and predicted half $[T_1,T_2]$;\\
        //\textit{For the initial state encoder}:\\
        Generate the latent initial states $z_i^{0,e}$ for each agent according to Eqn~\ref{eq:sample};\\
       //\textit{For the environment encoder}:\\
        Compute the latent representation for exogenous factors as in Eqn~\ref{eqn:environment};\\
        //\textit{For the generative model}:\\
        Given the latent initial states, the latent exogenous factors, and timestamps to predict $[T_1,T_2]$, solve the ODE function in Eqn~\ref{eqn:ode};\\
        //\textit{For the decoder}:\\
        Compute predicted node dynamics based on the decoding likelihood $ p(\bm{y}_{i}^{t,e} | \bm{z}_{i}^{t,e})$;
    }
    Update the parameters $\phi$ and $\theta$ by optimizing loss term defined in Eqn.~\ref{eqn:loss_all};
}
\end{algorithm}